\documentclass[preprint,12pt]{elsarticle}
\usepackage{graphicx}% Include figure files
\usepackage{dcolumn}% Align table columns on decimal point
\usepackage{bm}% bold math
\usepackage{amsfonts}
\usepackage{setspace}
\usepackage{color}
\usepackage{epsfig}
\usepackage{times}
\usepackage{booktabs} 
\usepackage{epstopdf}
\usepackage{multirow}
\usepackage{float}
\usepackage{algorithm}
\usepackage{threeparttable}
\usepackage{algpseudocode}
\usepackage{algorithmicx}
\usepackage{ragged2e}
\usepackage{ulem}
\usepackage{orcidlink}
\usepackage{hyperref}
\usepackage{soul}
\usepackage[T2A]{fontenc}
\usepackage[utf8]{inputenc}
\usepackage[greek,russian,spanish,english]{babel}
\usepackage[T1]{fontenc}

\usepackage{amsmath}

\usepackage[title]{appendix}
\usepackage{threeparttablex}

 \usepackage{colortbl}  %彩色表格需要加载的宏包
\usepackage{array} 

\usepackage{ulem} %by Fan

\usepackage[section]{placeins} 

\usepackage{cleveref}

\usepackage{amssymb}

\journal{ }

\begin{document}

\begin{frontmatter}

\title{Comprehensive Analysis of Network Robustness Evaluation Based on Convolutional Neural Networks with Spatial Pyramid Pooling}

\author[fn1]{Wenjun Jiang~\orcidlink{0000-0002-2244-5196}}

\author[fn2]{Tianlong Fan\corref{cor1}~\orcidlink{0000-0002-9456-6819}}
\ead{tianlong.fan@ustc.edu.cn}

\author[fn1]{Changhao Li}

\author[fn1]{Chuanfu Zhang\corref{cor1}}
\ead{zhangchf9@mail.sysu.edu.cn}

\author[fn1]{Tao Zhang}

\author[fn1]{Zong-fu Luo\corref{cor1}~\orcidlink{0000-0003-4871-4216}}
\ead{luozf@mail.sysu.edu.cn}

\cortext[cor1]{Corresponding author}

\affiliation[fn1]{organization={School of Systems Science and Engineering, Sun Yat-sen University},
city={Guangzhou},
postcode={510275},
country={China}}

\affiliation[fn2]{organization={School of Cyber Science and Technology, University of Science and Technology of China},
city={Hefei},
postcode={230026},
country={China}}

\begin{abstract}

Connectivity robustness, crucial for network understanding, optimization, and repair, has been evaluated traditionally through time-consuming and often impractical simulations. Fortunately, machine learning provides a novel solution. However, unresolved challenges persist: performance in more general edge removal scenarios, capturing robustness via attack curves instead of directly training for robustness, scalability of predictive tasks, and transferability of predictive capabilities. Here, we try to address these challenges by designing a convolutional neural networks (CNN) model with spatial pyramid pooling networks (SPP-net), adapting existing evaluation metrics, redesigning the attack modes, introducing appropriate filtering rules, and incorporating the value of robustness as training data. Results indicate that the CNN framework consistently provides accurate evaluations of attack curves and robustness values across all removal scenarios when the evaluation task aligns with the trained network type. This effectiveness is observed for various network types, failure component types, and failure scenarios, highlighting the scalability in task scale and the transferability in performance of our model. However, the performance of the CNN framework falls short of expectations in various removal scenarios when the predicted task corresponds to a different network type than the one it was trained on, except for random node failures. Furthermore, our work suggests that directly predicting robustness values yields higher accuracy than capturing them through attack curve prediction. In addition, the observed scenario-sensitivity has been overlooked, and the transferability of predictive capability has been overestimated in the evaluation of network features in previous studies, necessitating further optimization. Finally, we discuss several important unresolved questions.
\end{abstract}


\end{frontmatter}

\section{Introduction}
The robustness of networks pertains to their capacity to uphold structural integrity in the face of component failures or external damage. In this context, maintaining a certain level of structural integrity is deemed essential for the network to sustain its normal operations. The surging demand for connection and the rapid evolution of communication technologies has expedited the proliferation of myriad networked systems. Robustness emerges as a pivotal engineering challenge in the design and maintenance of various systems, especially in infrastructure and military networks. Typically, component failures, such as node \cite{cohen2000resilience, li2019robustness} or edge \cite{estrada2006network, zeng2012enhancing} failures, tend to transpire randomly \cite{albert2000error, fan2021rise}. Nonetheless, these failures can lead to shifts in load, setting off extensive cascading repercussions \cite{wang2008attack,wen2020overview, zhang2023analysis}. Moreover, systems frequently confront highly disruptive malicious attacks \cite{li2023adaptive} from external sources, which can result in unforeseeable losses or even catastrophic events \cite{buldyrev2010catastrophic}. These factors underscore the importance of research into robustness. Typically, during the process of failure or attack, a sequence of values, indicative of a specific network attribute, is used to assess network robustness.

It is worth noting that the study of network dismantling and robustness is inherently interconnected. The former is focused on swiftly disrupting a network's structure or paralyzing its functionality, while the latter emphasizes the network's ability to tolerate disruptions and maintain its operation, alongside strategies for enhancing this capacity \cite{valente2004two,xu2023assessing}. In both categories of research, a fundamental inquiry arises: how can we accurately evaluate network robustness at various stages? Specifically, which sequence of metrics best encapsulates a network's robustness? To address this question, several methods have emerged, which can be categorized into four classes:
\begin{itemize}
\item[1)]
Evaluation based on Network Topological Statistics: This category includes metrics such as connectivity measures \cite{frank1970analysis,1968On,liu2017comparative}, average path length \cite{albert2000error,cohen2001breakdown}, average geodesic length \cite{holme2002attack}, the size of the largest connected component (LCC) \cite{zeng2012enhancing,albert2000error,holme2002attack,lordan2019exact}, indicators considering the degree of cyclicity in a network \cite{fan2021characterizing}, and other related parameters \cite{wang2014damage}. However, these methods often suffer from high computational costs or infeasibility, particularly when dealing with large-scale and dynamic networks.
\end{itemize}
\begin{itemize}
    \item [2)]
Methods based on Percolation Theory: These methods are grounded in percolation theory applied to random graphs \cite{cohen2000resilience,holme2002attack,schneider2011mitigation}. They provide the critical node fraction that must be removed for network collapse. Initially limited to networks with Poisson distributions and random failures, they have been extended to encompass more general networks \cite{cohen2000resilience,cohen2001breakdown,paul2005resilience,callaway2000network,cai2020robustness} and malicious attacks \cite{cohen2001breakdown}. However, these methods primarily focus on the critical state of network collapse, assuming that the network has already sustained significant damage. Consequently, they do not fully reflect network robustness in alternative scenarios. Furthermore, many networks lack a critical state \cite{cohen2000resilience,schneider2011mitigation}.
\end{itemize}
\begin{itemize}
    \item [3)]
Measurements based on Matrix Spectra: This class comprises measurements based on matrix spectra \cite{yamashita2020predictability,wu2011spectral}, such as spectral radius \cite{jamakovic2006robustness}, spectral gap \cite{chan2016optimizing}, and natural connectivity \cite{wu2011spectral} derived from the adjacency matrix, as well as algebraic connectivity \cite{jamakovic2008robustness} and effective resistance \cite{chan2016optimizing}, which are based on the Laplacian matrix. Spectral measurements are straightforward to compute, but their relationship with network robustness is still not well-established. Additionally, evaluating their performance often depends on the LCC as a benchmark, and some measurements cannot assess the robustness of disconnected networks or scenarios involving edge removal \cite{zeng2012enhancing}.

\end{itemize}
\begin{itemize}
    \item [4)]
Metrics Incorporating Additional Network Properties: Metrics in this category consider network properties related to network flow \cite{cai2021network,si2022measuring}. However, these metrics typically necessitate knowledge of individual edge capacities or other additional information, which may not be applicable to many real-world networks.
\end{itemize}

Methods in the first category often fail to capture critical information due to the redundancy of trivial information, except for the LCC size, which directly mirrors the scale of the network's main body that maintains normal functionality and is the most widely employed metric. However, determining the LCC size involves a series of simulations, rendering the process computationally intensive and, in some cases, impractical, particularly for rapidly expanding or dynamic networks. Fortunately, the emergence of machine learning methods based on convolutional neural networks (CNN) has opened a new pathway for addressing this challenge by predicting the sequence of LCC sizes (attack curves) for robustness assessment. The adoption of CNNs as a promising future trend can be attributed to three key advantages: 1) Performance advantage: CNNs have demonstrated exceptional performance in image processing \cite{karpathy2014large} and are a superior deep learning architecture. The network's adjacency matrix can be directly treated as an image; 2) Efficiency advantage: once the training of the CNN is completed, the evaluation of robustness can be achieved instantaneously; 3) Generalization advantage: network topologies are inherently complex, with significant variations among different networks. The relationship between the order of node or edge removal and the rate of LCC decrease remains elusive. However, CNNs, equipped with robust fault-tolerance and self-learning capabilities, excel in handling such scenarios. They do not require an in-depth understanding of the specific network background, formation mechanisms, or functionalities. Furthermore, CNNs exhibit significantly superior generalization capabilities compared to other machine learning methods.

In this regard, pioneers have made attempts to cope with these challenges. The artificial neural network (ANN), a fundamental neural network model, was initially used to predict the controllability robustness of networks \cite{dhiman2021using}. Subsequently, the CNN, recognized for its excellence in image processing, was independently harnessed to predict both controllability robustness \cite{lou2020predicting} and connectivity robustness (referred to as `robustness' hereafter) within a unified framework \cite{lou2021convolutional}. The accuracy of controllability robustness prediction saw further enhancements by transitioning from a single CNN to multiple CNNs \cite{lou2021knowledge}. Furthermore, incorporating prior knowledge for network classification before prediction \cite{wu2022predicting}, or applying network filtering both before and after prediction \cite{lou2021knowledge, lou2023classification}, has shown potential to further refine prediction accuracy. Beyond network robustness, machine learning-based methods have also been instrumental in discovering more effective network dismantling strategies \cite{fan2020finding}. Additionally, there have been recent comprehensive reviews \cite{freitas2022graph,lou2023structural} summarizing the relevant findings and advancements in such studies, underscoring the remarkable effectiveness and efficiency of machine learning methods.

Although employing machine learning frameworks to address the conventional problem of network robustness evaluation holds potential, several pressing issues demand immediate attention. First, node removal represents a specific case of edge removal, essentially equivalent to removing all edges from a node at each step. It's imperative to explore whether machine learning frameworks can accurately predict network robustness in the context of edge removal. Second, the scalability of machine learning models remains a significant roadblock. Despite previous attempts at randomly reducing the network in advance \cite{lou2021knowledge} or dimensionality reduction through feature extraction \cite{lou2022learning}, these methods have only produced limited effects and introduced uncontrollable information loss. Third, current methodologies primarily aim to enhance algorithm performance by minimizing the disparity between the predicted attack curve and the simulated attack curve (constructed through the relative size of the LCC at each step). However, the ultimate network robustness is expressed as a scalar value, Robustness, representing the average of the relative sizes of the LCC at each step. In practice, although each simulated curve captures the associated robustness value, a crucial parameter, this aspect has been overlooked during prior training efforts.

This paper tries to tackle these issues by employing a CNNs model with SPP-net \cite{he2015spatial}, designing new simulated removal methods for differently sized training networks, and training the robustness value of networks as a key input along with the attack curve. Extensive testing of the proposed CNN framework across four distinct removal scenarios underscores its remarkable timeliness. However, the performance reveals its dependency on both specific scenarios and training data, in contrast to the overly optimistic expectations of earlier single-scenario-focused studies. The paper highlights the remarkable facets of the CNN framework, conducts a thorough analysis to elucidate potential contributing factors to its limitations, and proposes enhancements to ameliorate its overall performance.

The remaining content is organized as follows. Section \ref{Sec2} introduces the problem of network robustness. Section \ref{Sec3} presents the CNN model with SPP-net employed in this study, along with the newly designed attack methods that exhibit scalability with respect to network size. Section \ref{Sec4} delineates the training intricacies of the CNN model. Section \ref{Sec5} showcases our results and analyses. Finally, Section \ref{Sec6} summarizes this study and raises several important existing issues related to network robustness evaluation and machine learning frameworks for future works.

\section{Robustness} \label{Sec2}

For the aforementioned reasons, we select the LCC as the evaluation metric for the connectivity robustness of networks in the event of node failures or attacks \cite{schneider2011mitigation}, denoted as $R_{n}$. To meet the uniform size requirement of the training data for the CNN model, the revised $R_{n}$ is defined as:
\begin{equation} \label{eq:RDN}
R_{n}=\frac{1}{T}\sum_{p = 0}^{(T-1)/T}G_{n}(p),
\end{equation}
where $p$ is the proportion of nodes removed from the network and ranges from 0 to $(T-1)/T$ in increments of $1/T$, $T$ represents the total number of different values for $p$. $G_{n}(p)$ is the relative size of the LCC after removing a proportion $p$ of nodes. The normalization factor $1/T$ allows for the comparison of networks of different sizes.

In the real world, edge failures or attacks are a more general case \cite{zeng2012enhancing}. In this case, the corresponding revised network connectivity metric, denoted as $R_{e}$, is defined as:
\begin{equation}\label{eq:RDl}
R_{e} =\frac{1}{S}\sum_{p = 0}^{(S-1)/S}G_{e}(p),
\end{equation}
where $p$ is the proportion of edges removed, $S$ represents the total number of different values for $p$, $G_{e}(p)$ is the relative size of the LCC after removing a proportion $p$ of edges.

In the scenario of random failure, each iteration involves randomly removing a proportion of $p$ nodes or edges from the original network, the relative size of the LCC is then recorded. This process is referred to as random node failure (RNF) and random edge failure (REF). On the other hand, in the case of the malicious attack, the nodes with the highest degree or edges with the high-edge-degree, calculated based on the original network, are removed from the original network at each step. This strategy is known as the high-degree adaptive attack (HDAA) \cite{morone2015influence} and high-edge-degree adaptive attack (HEDAA) \cite{holme2002attack}. Additionally, the edge degree $k_e$ of an edge $e$ is defined as $k_e = k_v \times k_w$, where $k_v$ and $k_w$ are the degrees of the two end nodes of edge $e$, respectively. It is noteworthy that each removal in this study is performed on the original network to maintain consistent label scales across networks of varying sizes and sparsity levels.

\section{CNN Model} \label{Sec3}

Complex networks offer a new dimension of extension and development for machine learning, wherein the training results possess inherent physical interpretations. This is made possible by leveraging network characteristics as the foundation for data representation, enabling efficient capture of topological space relationships. Consequently, machine learning techniques establish meaningful connections within complex networks. As a result, the resolution of problems involving large-scale data becomes more feasible and convenient.

\begin{figure} [t]
\centering
\includegraphics[width=\linewidth]{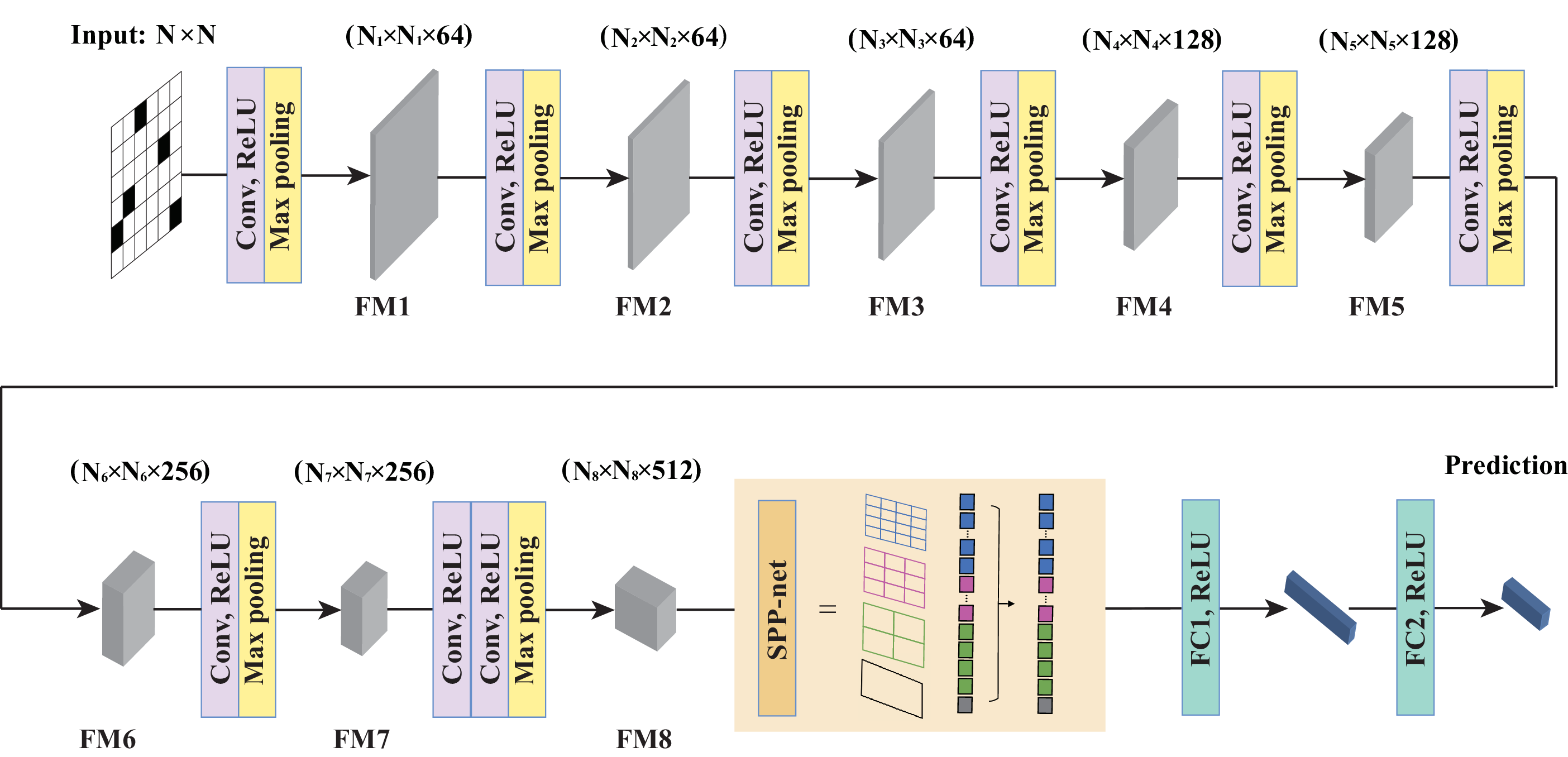}
\caption{[color online, 2-column] The architecture of the CNN employed in this study comprises four main components, namely the input layer, eight sets of convolutional layers, SPP-net, and two fully connected layers (FC). Each block within these convolutional sets typically includes one or two convolutional layers, a rectified linear unit (ReLU) activation function, and a max pooling layer. The size of the feature map (FM) is denoted as $N_i= \lceil \frac{N}{2^{i}} \rceil$, for $i=1,2,...,8$, where $N$ represents the number of nodes in the input network.}
        \label{fig:cnn}
\end{figure}

CNNs are specifically designed to process grid-structured data, such as binary image data represented as a two-dimensional grid of pixels. CNNs integrate the feature extraction function into a multilayer perceptron through structural reorganization and weight reduction, omitting the complex feature map extraction process before recognition \cite{simonyan2014very}.

The proposed CNN architecture, illustrated in Figure ~\ref{fig:cnn}, primarily consists of four components: the input layer, eight sets of convolutional layers, SPP-net and two fully connected layers (FC). Specifically, the first stage involves the input layer, which receives $N \times N$ images as inputs. Subsequently, the convolutional layer systematically extracts local features through the application of a diverse set of convolutional kernels, producing feature maps (FMs) as outputs. The specific parameters of the CNNs are detailed in Table ~\ref{tab_1}, and the use of a small stride in the convolution operation helps preserve detailed information within the network.

After the convolution operation, an activation function is typically applied, leveraging its nonlinearity to enable the neural network to approximate almost any prescribed nonlinear function. An example of a widely adopted activation function is the rectified linear unit (ReLU) \cite{nair2010rectified}, which directly outputs the input value if nonnegative; otherwise, it returns zero. The pooling layer, which serves to reduce the number of features and parameters, comes in two forms: average pooling and maximum pooling. In this context, the pooling layer utilizes the max pooling method with a stride of 2, resulting in a reduction of the feature map size exclusively within the pooling layer.

The fully connected layer (FC) requires a fixed input vector dimension, necessitating a flattened vector of predetermined size. This leads to maintaining a consistent network scale for both training and prediction. To tackle this challenge, SPP-net \cite{he2015spatial} was introduced. SPP-net addresses the difficulty of accommodating diverse input image sizes. In contrast to traditional CNNs, which require structural adjustments for varying input sizes, SPP-net enhances the network's robustness. This is achieved by partitioning the input feature map into grids at different scales and applying pooling operations on each grid. The pooled results from all scales are subsequently concatenated, thereby preserving spatial information from the input image while accommodating a range of input sizes.

\begin{table}[t]
\caption{Parameters of the proposed CNN framework}
\centering
\begin{tabular}{ccccc}
\toprule
\textbf{Group}                   & \textbf{Layer}     & \textbf{Kernel} & \textbf{Stride} & \begin{tabular}[c]{@{}c@{}}\textbf{Output}\\ \textbf{channel}\end{tabular} \\ 
\midrule
\multirow{2}{*}{Group1} & Conv3-64  & 3×3    & 1      & 64                                                       \\
                        & Maxpool   & 2×2    & 2      & 64                                                       \\
\multirow{2}{*}{Group2} & Conv3-64  & 3×3    & 1      & 64                                                       \\
                        & Maxpool   & 2×2    & 2      & 64                                                       \\
\multirow{2}{*}{Group3} & Conv5-64  & 5×5    & 1      & 64                                                       \\
                        & Maxpool   & 2×2    & 2      & 64                                                       \\
\multirow{2}{*}{Group4} & Conv3-128 & 3×3    & 1      & 128                                                      \\
                        & Maxpool   & 2×2    & 2      & 128                                                      \\
\multirow{2}{*}{Group5} & Conv3-128 & 3×3    & 1      & 128                                                      \\
                        & Maxpool   & 2×2    & 2      & 128                                                      \\
\multirow{2}{*}{Group6} & Conv3-256 & 3×3    & 1      & 256                                                      \\
                        & Maxpool   & 2×2    & 2      & 256                                                      \\
\multirow{2}{*}{Group7} & Conv3-256 & 3×3    & 1      & 256                                                      \\
                        & Maxpool   & 2×2    & 2      & 256                                                      \\
\multirow{3}{*}{Group8} & Conv3-512 & 3×3    & 1      & 512                                                      \\
                        & Conv3-512 & 3×3    & 1      & 512                                                      \\
                        & Maxpool   & 2×2    & 2      & 512                                                      \\ \hline
\end{tabular}
\label{tab_1}
\end{table}

Consider a feature map $X$ as the input to SPP-net after a series of convolution operations. Let $D$ represent the number of output channels in the last layer, and $H$ be the total number of bins in the spatial pyramid. The spatial pyramid pooling (SPP) process is mathematically defined as follows:
\begin{equation}\label{eq:spp}
V = [P_1(X_1), P_2(X_2), P_3(X_3), P_4(X_4)],
\end{equation}
where $P_i$ denotes the pooling operation at level $i$ and $X_i$ is the pooled feature map at that level. $D\times H$ is the final vector that the fully connected layer needs. For our SPP-net, the pyramid consists of four levels: $\{ 4 \times 4$, $3 \times 3$, $2 \times 2$, $1 \times 1 \}$, thus $H=4\times4+3\times3+2 \times 2+1 \times 1=30$ and $D = 512$ as indicated in Table \ref{tab_1}. This configuration results in a total of 30 bins, producing fixed-length vectors of $30 \times 512 = 15,360$. Following the SPP process, these vectors are then input into fully connected layers to facilitate the achievement of the learning objectives.

The choice of the loss function is crucial in guiding effective model learning. Here, the mean squared error (MSE) between the predicted vector (referred to as an attack curve and a scalar, Robustness), and the simulated vector (same as above) is employed as the loss function, as follows:
\begin{equation}\label{eq:L2loss}
J_{MSE}=\frac{1}{L}\sum_{i=1}^{L+1} \parallel y_i-\hat{y}_i \parallel^2,
\end{equation}
where $\parallel \cdot \parallel$ is the Euclidean norm, $y_i$ and $\hat{y}_i$ is the $i$th element of the predicted vector given by the CNN and the simulated vector respectively. The first $L$ elements are the elements of the attack curve, while the $(L+1)$th element represents the value of Robustness. The objective of training the model is to minimize the $J_{MSE}$ as much as possible, with its minimum value of 0 indicating that the predicted results are exactly identical to the simulation. Generally, an undirected and unweighted network with $N$ nodes is represented as an adjacency matrix, which can be directly regarded as a binary image. In this representation, a value of 1 in an element signifies the presence of an edge, while 0 signifies the absence of an edge.

It is necessary to introduce a previous similar model, CNN-RP model \cite{lou2021convolutional}, and use it as a baseline for the proposed model. CNN-RP is a basic CNN model used to predict the network's connectivity robustness. The differences between CNN-RP and our model are as follows: 1) The architecture of our model is shown in Figure \ref{fig:cnn} and includes smaller convolution kernels, while the CNN-RP model consists of seven convolutional layers and two fully connected layers. 2) The CNN-RP model's simulation data is obtained by sequentially removing nodes, which, due to the constraints of the CNN regarding the scale of training data and the prediction task, limits its capability to handle tasks strictly of the same scale as the training data, lacking scalability. 3) The CNN-RP model can only handle robustness prediction in node removal scenarios. 4) In addition to boundary value filters, the CNN-RP model also includes a monotonic non-decreasing constraint. If the prediction results do not satisfy this condition, outliers are replaced by interpolated values generated according to the mean. Furthermore, for comparability, we replaced the original robustness calculation way of the CNN-RP model with Equation \ref{eq:RDN}, noting that this does not affect the model's performance.

\section{Model Training}\label{Sec4}

The initial part in Figure ~\ref{fig:cnn} depicts that the obtained image is input into the CNN for training. Having SPP-net integrated into the CNN model enhances its ability to handle networks of different sizes for prediction. However, the training data still imposes a restriction of having at most two different sizes (single-size or multi-size) \cite{he2015spatial}. For simplicity, we select the mode of single-size for the training phase, namely a fixed network size, 1,000, and also $T=S=1,000$. Synthetic networks are employed for training purposes. For each training network, the model takes an input in the form of a binary tuple $(G, L_{RS})$, where $G$ represents the adjacency matrix of the network, and $L_{RS}$ is a label vector with a dimension of 1,001. The first 1,000 elements of $L_{RS}$ correspond to the relative size of the LCC after removing the corresponding proportion of nodes or edges, also known as the attack curve. The last element represents the network's robustness, denoted as $R_n$ or $R_e$, representing the normalized values of the area under the attack curve.

Synthetic networks used for training include ER model networks \cite{1968On} and BA model networks \cite{barabasi1999emergence}, which serve as representatives of assortative and disassortative networks respectively, each with three different average degrees $\langle k \rangle=4$, 6 and 8 respectively. Consider random failures as an example, each attack involves randomly removing a proportion $p$ of nodes or edges from the original network, where $p$ ranges from 0 to 0.999 in increments of 0.001. Subsequently, the remaining network's $R_n$ or $R_e$ is computed. During the prediction phase, target networks of varying sizes, both synthetic and empirical, are fed into the model, and the returned result is a vector of length 1,001, which has the same value type as $L_{RS}$. Lastly, to address the issue of predicted values lacking physical meaning, such as when $G_{n}$ or $G_{e}$ is greater than 1 or less than 0, a filtering mechanism can be implemented to normalize them. Values greater than 1 are set to 1, and values less than 0 are set to 0. In this context, we refrain from introducing additional monotonicity filters for robustness assessment, aiming to elucidate the genuine predictive capacity of the CNN model.

To train the CNN model with SPP-net, 1,000 synthetic networks with different average degrees are generated for each removal scenario. Among these, 800 cases are allocated for training, 100 for cross-validation, and 100 for testing. For example, in the random edge removal scenario, there are a total of 2 (ER network and BA network) $\times$ 3 ($\langle k \rangle=$ 4, 6, 8) $\times 800 = 4,800$ training instances, 600 cross-validation instances, and 600 test instances. The number of training epochs is set to 20, and the batch size is fixed at 4.

During training, the instances are shuffled, and Figure ~\ref{fig:loss} in Appendix shows the loss values for different attack scenarios. The training loss gauges how well the model performs on the training data by minimizing the associated loss function during parameter adjustments. Simultaneously, the validation loss assesses the model's ability to generalize to unseen data, acting as a check against overfitting. Monitoring the gap between training and validation losses helps identify potential overfitting issues. Balancing these losses assists in selecting models that exhibit superior performance when confronted with new data. The values of train loss and validation loss are both small and converge as the epoch increases. This indicates that the model's performance on training and validation data is becoming consistent. Such a scenario typically suggests that the model possesses good generalization capabilities, performing well on unseen data. The gradual convergence of training and validation losses may imply that the model is gradually overcoming overfitting during the learning process, adapting better to diverse data distributions. Observing this trend is often a sign of stable performance improvement throughout the training process.

The experiments are performed on a PC with a 64-bit Windows 11 Operating System, installed with an Intel (R) 16-Core i7-11700F (2.50GHz) CPU.

\begin{figure} [t]
\centering
\centerline{\includegraphics[width=\linewidth]{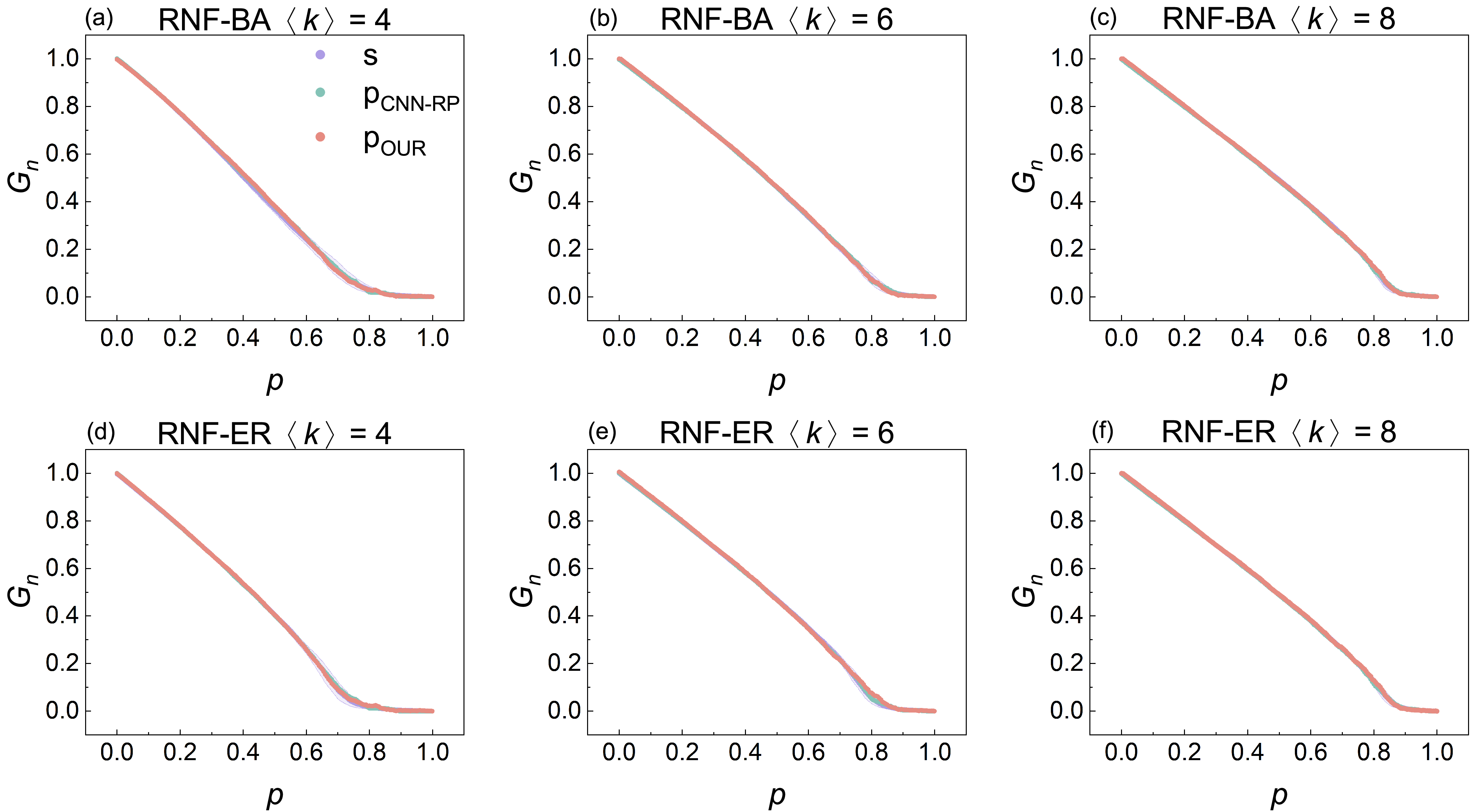}}
\caption{[color online, 2-column] Attack curves on synthetic networks under random node failure (RNF). Each panel's title specifies the type of failure, network model, and average degree. In this context, $p$ denotes the proportion of removed nodes, and $G_n$ represents the relative size of the LCC. The curves $p_\text{CNN-RP}$ and $p_\text{OUR}$ illustrate the predicted attack curves, while $s$ represents the simulated attack curves. Light shading is utilized to depict the standard deviation. Each dot is the average of 100 networks from the test set.}
\label{fig:RNF}
\end{figure}

\section{Results} \label{Sec5}

We conducted evaluation experiments considering four removal scenarios: random node failure (RNF), malicious node attack with HDAA, random edge failure (REF) and malicious edge attack with HEDAA, covering both synthetic networks and empirical networks. The synthetic networks consisted of ER networks and BA networks and the empirical networks are diverse in field. These experiments comprehensively demonstrate the holistic evaluation capabilities of the CNN model with SPP-net across all of these scenarios.

\subsection{Synthetic Networks}
 
Figure ~\ref{fig:RNF} depicts the attack curves of synthetic networks under random node failure. It demonstrates a remarkable fit between the predicted and simulated results, highlighting the exceptional predictive ability shared by our model and CNN-RP. As the average degree increases, the area under the attack curve becomes larger, indicating stronger network robustness. In addition, as shown in Table ~\ref{tab:A_RNF} in Appendix, with the increase in average degree, the mean standard deviations ($\Bar{e}_{sim}$, $\Bar{e}_{our}$, and $\Bar{e}_{CNN-RP}$) for the three methods are decreasing. This indicates a reduction in their variability, implying greater stability in both the simulation and prediction results. Remarkably, the mean standard deviations of both predicted methods are lower than that of the simulated method, and ours is even lower than the baseline method. Table ~\ref{tab:A_RNF} also presents the mean difference between simulation and prediction results, with ours being lower than the baseline. These findings suggest that as network density increases, CNN predictions become more accurate, and our model demonstrates greater accuracy and stability than CNN-RP.

\begin{figure} [t]
\centering
\centerline{\includegraphics[width=\linewidth]{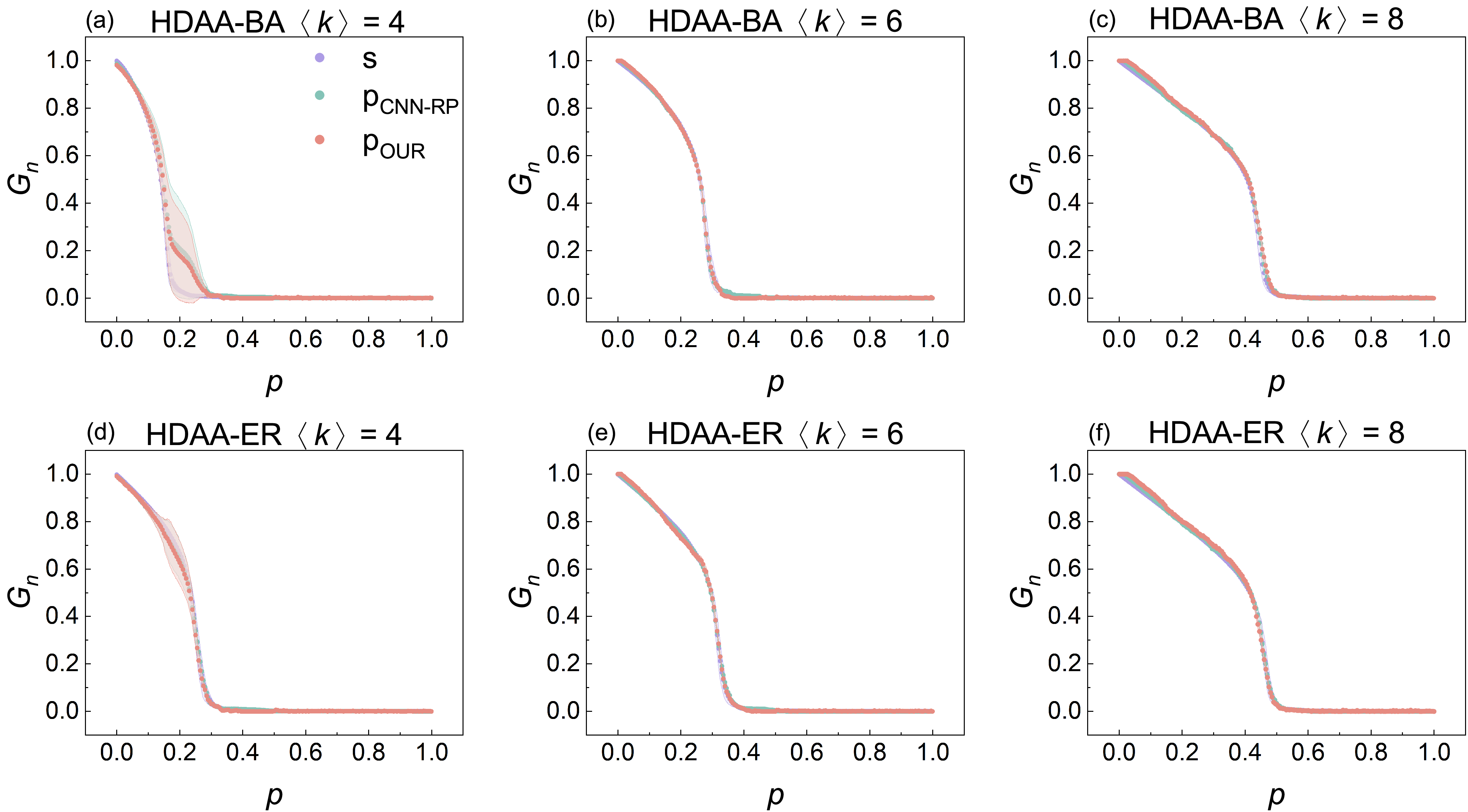}}
\caption{[color online, 2-column] Attack curves on synthetic networks under high-degree adaptive attack (HDAA). Each panel's title specifies the type of failure, network model, and average degree. In this context, $p$ denotes the proportion of removed nodes, and $G_n$ represents the relative size of the LCC. The curves $p_\text{CNN-RP}$ and $p_\text{OUR}$ illustrate the predicted attack curves, while $s$ represents the simulated attack curves. Light shading is utilized to depict the standard deviation. Each dot is the average of 100 networks from the test set.}
\label{fig:HDAA}
\end{figure}

Figure ~\ref{fig:HDAA} illustrates the results of high-degree adaptive attacks targeting nodes. In the initial phase characterized by a rapid decline in the attack curves, the predictions do not exhibit a perfect match with the simulations; rather, they display noticeable fluctuations. Nevertheless, as the network becomes denser, these two curves converge almost completely, with the stability of the prediction outcomes quickly improving. This is clearly reflected in Table ~\ref{tab:A_HDAA} in Appendix, where the mean standard deviations of the two prediction results rapidly decrease to below those of the simulated results. Regarding the mean difference from the simulated results, there is no significant difference between the two prediction methods.

We also present for the first time the prediction results of edge removal in two scenarios: Figure ~\ref{fig:REF} and Table ~\ref{tab:A_REF} in Appendix for random edge failures, and Figure ~\ref{fig:HEDAA} and Table ~\ref{tab:A_HEDAA} in Appendix for high-edge-degree adaptive attacks. In the context of random edge failures, our model exhibits outstanding performance, similar to the node removal scenario, as shown in Figure ~\ref{fig:REF}. The mean standard deviations of the predicted values also remain lower, and the mean difference between them is at a similar level as in the previous results, as shown in Table ~\ref{tab:A_REF}. However, this is not the case for malicious edge attacks, especially in the phases before and after rapid network connectivity decline, as shown in and Figure ~\ref{fig:HEDAA}. Meanwhile, the mean standard deviations of the prediction are higher than those from simulation, and the mean difference between them is also higher than in the previous results, as shown in Table ~\ref{tab:A_HEDAA}. Although the gap between the simulated and predicted curves decreases as the network density increases, corresponding to an improvement in the accuracy of Robustness value prediction, the fluctuations in the prediction results remain non-negligible. This outcome suggests that the predictive capability of machine learning frameworks for network topology changes in more general edge removal scenarios is not as optimistic as prior studies that focused on node removal.

\begin{figure} [t]
\centering
\centerline{\includegraphics[width=\linewidth]{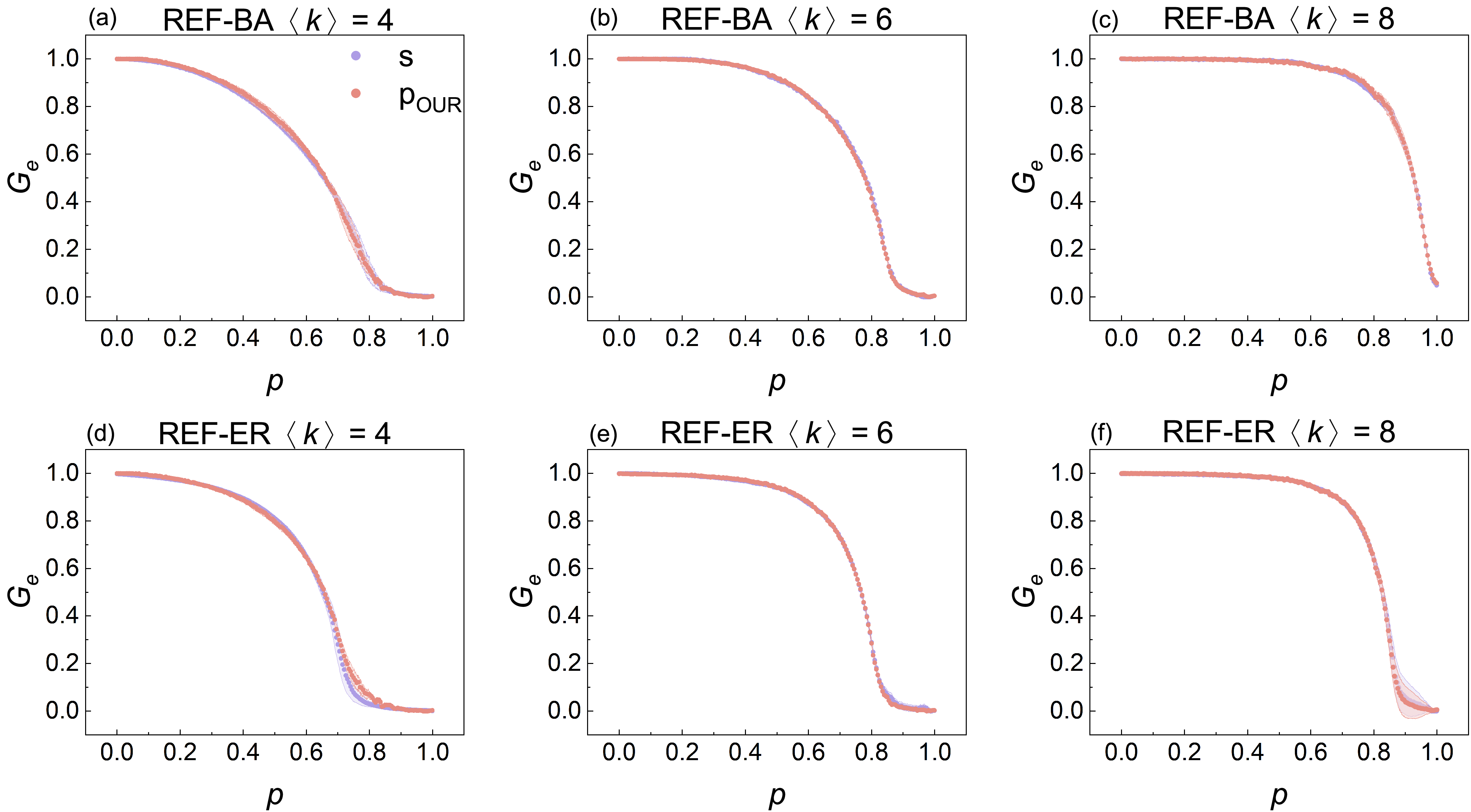}}
\caption{[color online, 2-column] Attack curves on synthetic networks under random edge failure (REF). Each panel's title specifies the type of failure, network model, and average degree. In this context, $p$ denotes the proportion of removed nodes, and $G_e$ represents the relative size of the LCC. The curve $s$ represents the simulated attack curve, while $p_\text{OUR}$ illustrates the predicted attack curve. Light shading is utilized to depict the standard deviation. Each dot is the average of 100 networks from the test set.}
\label{fig:REF}
\end{figure}

\begin{figure} [t]
\centering
\centerline{\includegraphics[width=\linewidth]{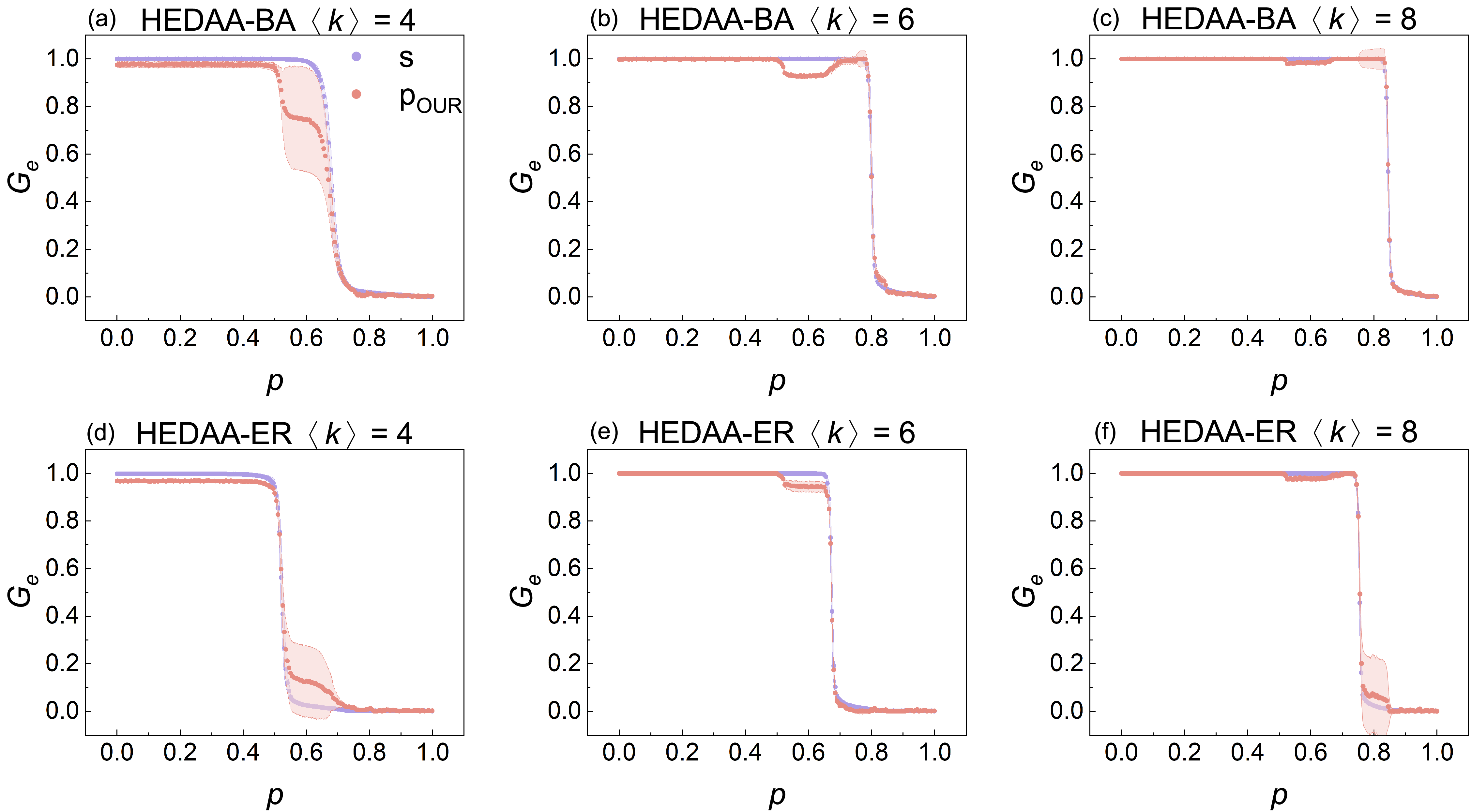}}
\caption{[color online, 2-column] Attack curves on synthetic networks under high-edge-degree adaptive attack (HEDAA). Each panel's title specifies the type of failure, network model, and average degree. In this context, $p$ denotes the proportion of removed nodes, and $G_e$ represents the relative size of the LCC. The curve $s$ represents the simulated attack curve, while $p_\text{OUR}$ illustrates the predicted attack curve. Light shading is utilized to depict the standard deviation. Each dot is the average of 100 networks from the test set.}
\label{fig:HEDAA}
\end{figure}

In terms of robustness, it can be observed that under malicious edge attacks, as shown in Figure ~\ref{fig:HEDAA}, BA networks exhibit stronger robustness compared to ER networks. Specifically, BA networks require a higher proportion of edge removals to reduce the relative size of the LCC to 0. This is mainly due to the presence of a few nodes with extremely high degrees in BA networks. These nodes contribute to high edge degrees for their adjacent edges. As a result, when the network is attacked using the HEDAA method, the removed edges are mostly concentrated locally, causing minimal disruption to the global connectivity of the network. On the other hand, concerning the predictive capabilities of the model, a comparison of all the aforementioned results reveals that the model's predictive capabilities for ER networks (homogeneous) and BA networks (heterogeneous) do not exhibit statistically significant differences, in contrast to the results for controllability robustness \cite{lou2020predicting}.

In comparing the predictive capabilities of the model across these four removal scenarios under the same training scale, several differences are evident. Firstly, random removals are more easily predicted, while malicious attacks are relatively more challenging to predict, whether involving nodes or edges. This is attributed to the heightened sensitivity of the LCC to removed nodes during the rapid descent phase in malicious attacks, where even slight discrepancies swiftly lead to significant inconsistencies. Secondly, edge removals are more challenging to predict than node removals, especially in the scenarios of malicious edge attacks. We attribute this, in addition to the previously mentioned factors, to a relatively insufficient level of training. Specifically, edge removals are essentially a more generalized form of node removals. In edge removals, the removal of two edges adjacent to the same node does not affect each other. Therefore, compared to node removals, edge removals have a greater number of possible combinations. Consequently, when the number of training networks is the same, the actual training depth for edge removals is much lower than that for node removals.

\begin{figure}[t]
\centering
\centerline{\includegraphics[width=\linewidth]{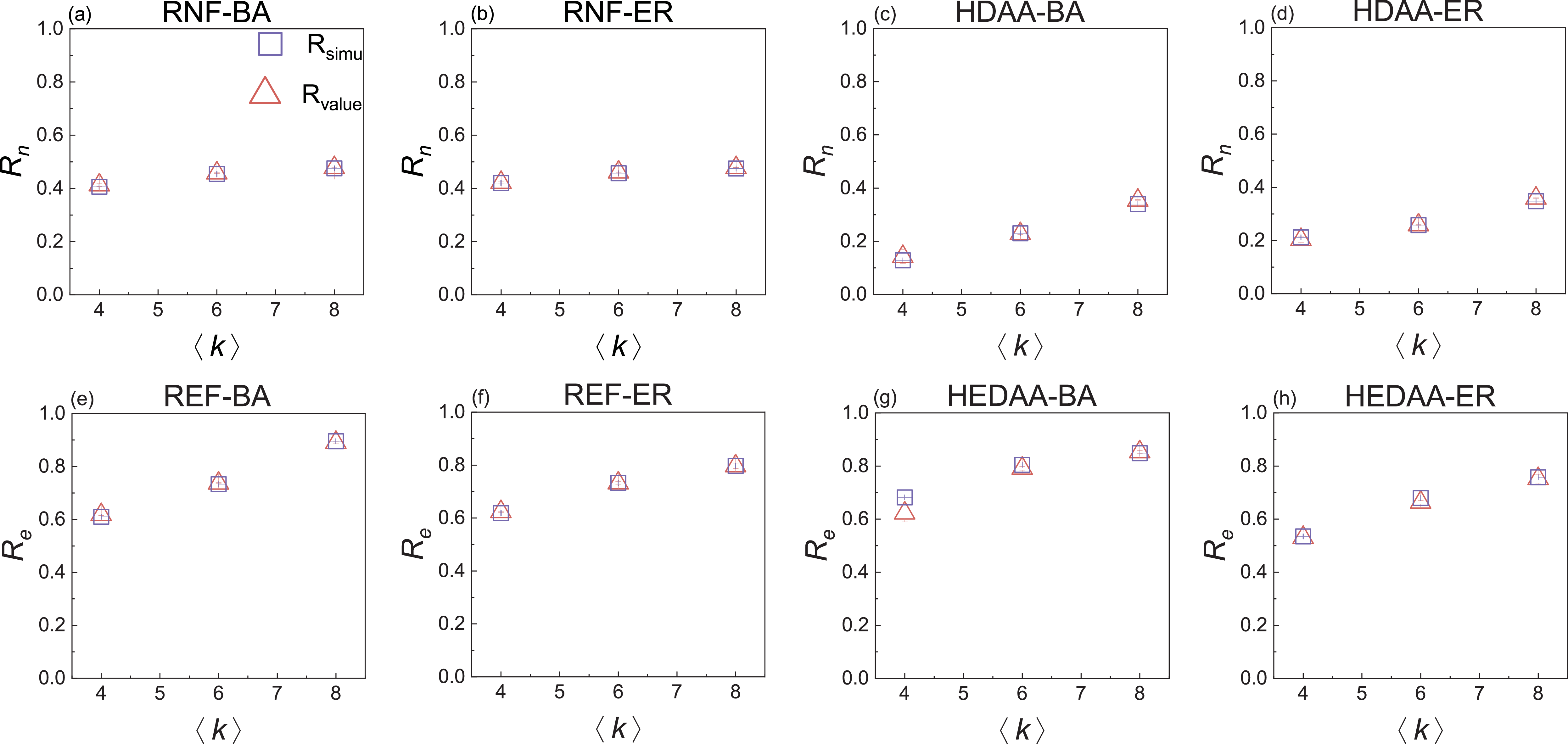}}
\caption{[color online, 2-column] Comparison of simulated and direct predicted robustness values ($R_n$ and $R_e$) across two synthetic networks and four removal scenarios. $R_\text{value}$ is the robustness value directly predicted by our method, while $R_\text{simu}$ is derived from the simulated attack curve. Here each signal is the average and error bar of 100 networks from the test set.}
\label{fig:R}
\end{figure}

In this study, we hypothesized that incorporating robustness value into the training data would be beneficial for predicting both attack curve and robustness value itself. This is validated in Figure ~\ref{fig:R}, which demonstrates the model's predictions of robustness values across two synthetic networks and four removal scenarios. As compared to the predictions of attack curves, such as in Figure ~\ref{fig:HEDAA}, the model exhibits superior performance when directly predicting robustness values, irrespective of the scenario. This is because, compared to predicting an attack curve (a high-dimensional value sequence), direct scalar value prediction is simpler and is not subject to the overestimation and underestimation issues associated with curve predictions during different stages.

\begin{table}[t]
\caption{Basic topological features of the three real-world networks. Here $N$ and $M$ are the number of nodes and edges respectively, $\langle k \rangle$ is the mean degree.}
\centering
\setlength{\tabcolsep}{5mm}{}
\begin{threeparttable}
\begin{tabular}{cccccc}
\hline
Network & $N$  & $M$  & $\langle k\rangle$  \\ \hline
USAirRouter     & 1226       & 2408       & 3.93   \\ 
HI-II-14        & 4165       & 13087       & 6.28     \\ 
Biological     & 9436       & 31182       & 6.546     \\ \hline
\end{tabular}
\label{tab:realnets}

\end{threeparttable}
\end{table}

\subsection{Empirical Networks}

In addition to predicting the attack curves and the direct robustness values of synthetic networks that are similar to the training data, we also demonstrate the performance of our model on untrained empirical network data with different scales. The results reflect the model's transferability in predictive capabilities and its scalability in handling different scales of prediction tasks.

\begin{figure}[t]
\centering
\includegraphics[width=\linewidth]{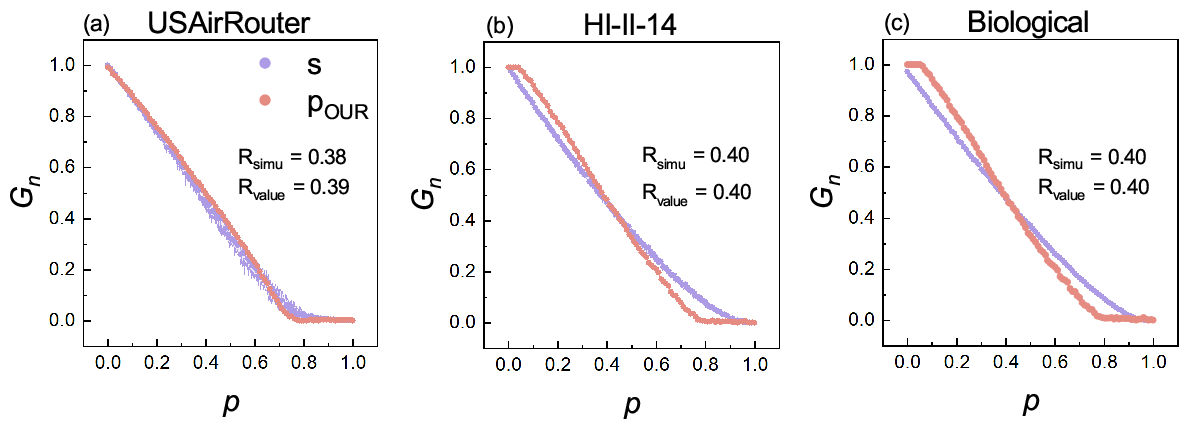}
\caption{[color online, 2-column] Attack curves and the robustness values on empirical networks under random node failure (RNF). Here $p$ and $G_n$ represent the proportion of removed nodes and the relative size of the LCC respectively. The curve $s$ represents the simulated attack curve, while $p_\text{OUR}$ illustrates the predicted attack curve. $R_\text{simu}$ is the robustness value derived from the simulated attack curve, while $R_\text{value}$ is the robustness value directly predicted by our method.}
\label{fig:real_Rn_RNF}
\end{figure}

We considered three empirical networks from disparate fields ~\cite{nr}: the USAirRouter network, HI-II-14 network (protein interaction network) and Biological network, with varying sizes and sparsity levels and differ significantly from the networks used for training the model. Table ~\ref{tab:realnets} provides the basic statistics for these networks, indicating significant disparities in both node and edge scale, as well as density. Figure ~\ref{fig:real_Rn_RNF} illustrates the predicted attack curves and direct robustness of the proposed model under the scenario of RNF. The predicted attack curves can be seen not to align well with the simulated attack curves; however, the directly predicted robustness values are either equivalent to or very close to the simulated results. This further underscores that direct robustness value prediction is a superior choice. In addition, The results highlight the excellent scalability of our model in handling diverse task scales and the transferability of its prediction capabilities.

Figure ~\ref{fig:empNet_3scenarios} in Appendix showcases the results for the remaining three attack scenarios on empirical networks, highlighting significant disparities between predicted and simulated curves. In comparison, the direct value predictions exhibit smaller differences when compared to the simulations. Specifically, in HDAA and REF scenarios, the predicted attack curves show a consistent declining trend and generally follow the variations of their simulated counterparts, albeit with considerable deviations between them. The differences in directly predicted robustness values, on the other hand, are relatively small, especially in the REF scenario for the Biological network. In the case of HEDAA, the predicted attack curve exhibits substantial fluctuations, which contradicts the expected monotonic behavior of robustness. The directly predicted robustness values also demonstrate significant discrepancies compared to the simulations. Overall, substantial room for improvement remains in the predictive capabilities for these three scenarios. Additionally, it is evident that random failures are more predictable than malicious attacks, and node removal is easier to predict than edge removal. 

The primary reason for these results is the lack of training based on empirical networks. Moreover, achieving the same level of training as node removal requires more training data for edge removal, attributed to the greater number of potential removal combinations for edges. Secondly, this might also stem from the increasing disparity between the edge scale in the removal scenario and the training data, resulting in a higher proportion of edges being removed for each fixed $p$.

Nevertheless, it is undeniable that the proposed CNN method thoroughly addresses the high computational time challenges of traditional simulations. Table ~\ref{tab:A_Time} in Appendix demonstrates the considerable advantages of the proposed CNN method in terms of evaluation time compared to traditional simulations. In any given scenario, evaluation based on the CNN method are either instantaneous or exhibit several orders of magnitude improvement, with some cases even showing a difference of up to 10,000 times faster.

\section{Conclusion and Discussion} \label{Sec6}

In this paper, we have provided a comprehensive overview of the mainstream types of network robustness evaluation metrics, highlighting their strengths and limitations. To tackle the challenges associated with network robustness evaluation in an effective and efficient manner, we have proposed a machine learning model as an alternative to traditional simulation methods. Our approach involves refining existing evaluation metrics to suit CNN models, introducing new simulated attack modes, employing a CNN with SPP-net, and refining the training data and the prediction targets. These measures collectively enhance evaluation accuracy, introduce scalability in task scale, and partially validate the transferability of performance across different contexts. Our findings suggest that the CNN framework for robustness assessment may not be as outstanding and consistent as previously suggested by one-sided results.

The extensive results demonstrate the excellent timeliness and the varying performance of the proposed CNN framework across different task scenarios. In prediction tasks that align with the training network type, the CNN exhibits outstanding predictive capabilities, irrespective of the type of component failure (nodes or edges), the removal scenarios (random failures and malicious attacks), or the specific evaluation task (predicting attack curve or robustness). For prediction tasks that deviate from the training network type, the CNN performs well in evaluating node random failure, showcasing its scalability and performance transferability in prediction tasks. However, its performance in the remaining three removal scenarios is subpar, which highlights an overlooked or incorrectly anticipated aspect in existing research. Our work underscores the potential of CNN-based models in assessing network properties and their applicability to a range of real-world scenarios, also suggesting possible avenues for further improving their performance.

Our research has also unveiled new questions and areas that require further improvement: 1) How can we optimize the prediction performance during the stages of drastic changes in network robustness? 2) Exploring methods to achieve scalable scale of training data is crucial, as it greatly impacts the applicability and performance of CNN models. 3) Designing more effective attack strategies based on machine learning techniques. 4) Can a single machine learning model handle both node removal and edge removal tasks for both random failures and malicious attacks?

Comprehensive comparisons indicate that, for these underperforming scenarios, thorough pre-training significantly enhances predictive performance. This involves increasing the number of training networks, diversifying the types of training networks, spanning the density range of training networks, and refining the granularity of single-step removals during training to reduce the number of nodes and edges encompassed in the removal proportion $p$. Moreover, the exploration of alternative deep learning approaches such as Graph Neural Networks (GNN) \cite{zhou2020graph} is deemed imperative for future endeavors.

Furthermore, in the realm of network robustness research itself, we need to address the question of whether it is possible and how to determine the worst-case scenario (lower bound) for the connectivity robustness of a given network, irrespective of the employed attack method.

The authors believe that these questions merit significant attention from the research community and should be pursued as future research objectives. Addressing these inquiries in a timely manner will undoubtedly lead to substantial breakthroughs and practical applications.

\section{Author contributions}%{CRediT authorship contribution statement}
Wenjun Jiang, Tianlong Fan and Zong-fu Luo conceived the idea and designed the study. Tianlong Fan, Zong-fu Luo, Chuanfu Zhang and Tao Zhang managed the study. Wenjun Jiang and Changhao Li collected and cleaned up the data. Wenjun Jiang, Tianlong Fan and Changhao Li performed the experiments under the leadership of Zong-fu Luo. Wenjun Jiang and Tianlong Fan wrote the manuscript, Changhao Li supplemented visualization. Zong-fu Luo, Tao Zhang and Chuanfu Zhang edited this manuscript. All authors discussed the results and reviewed the manuscript.

\section{Declaration of competing interest}
The authors declare no competing interests.

\section{Acknowledgments}
This work was supported in part by the National Natural Science Foundation of China (Grant No. T2293771), the National Laboratory of Space Intelligent Control (No. HTKJ2023KL502003), the Fundamental Research Funds for the Central Universities, Sun Yat-sen University (No. 23QNPY78) and the STI 2030--Major Projects (No. 2022ZD0211400).

\journal{Journal of \LaTeX\ Templates}
 \bibliographystyle{elsarticle-num} 
\bibliography{robustnesspreref.bib}

%\appendix
%\section{}
\newpage
\addcontentsline{toc}{section}{Appendix}
\section*{Appendix}\label{secA1}

\setcounter{table}{0}
\renewcommand\thetable{A\arabic{table}}  
\setcounter{figure}{0}
\renewcommand\thefigure{A\arabic{figure}}

\begin{table}[!htb]
%\begin{threeparttable}
\caption{Mean standard deviations of simulation and prediction and the mean difference between them under random node failure (RNF)}
    \centering
    \begin{tabular}{l|ccc|ccc}
        \hline
        \multirow{2}*{Error Types} &  \multicolumn{3}{c}{ER networks} & \multicolumn{3}{c}{BA networks} \\ 
        \cmidrule{2-7} 
        & $\langle k\rangle = 4$  &  $\langle k\rangle = 6$  &  $\langle k\rangle = 8$ & $\langle k\rangle = 4$  &  $\langle k\rangle = 6$  &  $\langle k\rangle = 8$ \\ 
        \hline
        $\Bar{e}_{sim}$ & 0.0106   & 0.0072     & 0.0046  & 0.0125    & 0.0075   &0.0044 \\
        $\Bar{e}_{our}$ & 0.0027   & 0.0007     & 0.0004  & 0.0048    & 0.0012   &0.0006 \\
        $\Bar{e}_{CNN-RP}$ & 0.0031   & 0.0020     & 0.0003  & 0.0048    & 0.0026   &0.0006 \\
        $\Bar{e}_{(sim,our)}$ & 0.0023   & 0.0055     & 0.0026  & 0.0053    & 0.0027   &0.0029 \\
        $\Bar{e}_{(sim,CNN-RP)}$ & 0.0030   & 0.0038     & 0.0039  & 0.0031    & 0.0036   &0.0046 \\  \hline
    \end{tabular}
%\TPTnoteSettings{\flushleft} % 将注释左对齐

%\begin{tablenotes}[htbp]
    %\item 
    \small \flushleft
    Note: $\Bar{e}_{sim}$, $\Bar{e}_{our}$ and $\Bar{e}_{CNN-RP}$ represent the mean standard deviation of 100 simulated $L_{RS}$, predicted $L_{RS}$ from our model and predicted $L_{RS}$ from CNN-RP. $\Bar{e}_{sim,our}$ is the difference between $p_i$ and $s_i$, $\Bar{e}_{sim,our} = \frac{\sum_{i=1}^L |p_i-s_i|}{L}$, where $p_i$ is the $i$th prediction from our model and $s_i$ is the $i$th simulation. $\Bar{e}_{(sim,our)}$ has the same function as $\Bar{e}_{(sim,CNN-RP)}$ but for CNN-RP model.
%\end{tablenotes}
\hrule
%\end{threeparttable}
\label{tab:A_RNF}

\end{table}

\begin{table}[!htb]
    \caption{
    Mean standard deviations of simulation and prediction and the mean difference between them under high-degree adaptive attack (HDAA)
    }
    \centering
    \begin{tabular}{l|ccc|ccc}
        \hline
        \multirow{2}*{Error Types} &  \multicolumn{3}{c}{ER networks} & \multicolumn{3}{c}{BA networks} \\ 
        \cmidrule{2-7} 
        & $\langle k\rangle = 4$  &  $\langle k\rangle = 6$  &  $\langle k\rangle = 8$ & $\langle k\rangle = 4$  &  $\langle k\rangle = 6$  &  $\langle k\rangle = 8$ \\ 
        \hline
        $\Bar{e}_{sim}$ & 0.0063   & 0.0061     & 0.0050  & 0.0050    & 0.0057   &0.0045 \\
        $\Bar{e}_{our}$ & 0.0131   & 0.0034     & 0.0024  & 0.0246    & 0.0038   &0.0028 \\
        $\Bar{e}_{CNN-RP}$ & 0.0121   & 0.0032     & 0.0034  & 0.0277    & 0.0027   &0.0036 \\
        $\Bar{e}_{(sim,our)}$ & 0.0078   & 0.0050     & 0.0096  & 0.0190    & 0.0040   &0.0088 \\ 
        $\Bar{e}_{(sim,CNN-RP)}$ & 0.0053   & 0.0053     & 0.0071  & 0.0067    & 0.0043   &0.0059 \\    \hline

    \end{tabular}

    \small \flushleft
    Note: $\Bar{e}_{sim}$, $\Bar{e}_{our}$ and $\Bar{e}_{CNN-RP}$ represent the mean standard deviation of 100 simulated $L_{RS}$, predicted $L_{RS}$ from our model and predicted $L_{RS}$ from CNN-RP. $\Bar{e}_{sim,our}$ is the difference between $p_i$ and $s_i$, $\Bar{e}_{sim,our} = \frac{\sum_{i=1}^L |p_i-s_i|}{L}$, where $p_i$ is the $i$th prediction from our model and $s_i$ is the $i$th simulation. $\Bar{e}_{(sim,our)}$ has the same function as $\Bar{e}_{(sim,CNN-RP)}$ but for CNN-RP model.

\hrule
    
\label{tab:A_HDAA}
\end{table}

\begin{table}[!htb]
    \caption{
    Mean standard deviations of simulation and prediction and the mean difference between them under random edge failure (REF)
    }
    \centering
    \begin{tabular}{l|ccc|ccc}
        \hline
        \multirow{2}*{Error Types} &  \multicolumn{3}{c}{ER networks} & \multicolumn{3}{c}{BA networks} \\ 
        \cmidrule{2-7} 
        & $\langle k\rangle = 4$  &  $\langle k\rangle = 6$  &  $\langle k\rangle = 8$ & $\langle k\rangle = 4$  &  $\langle k\rangle = 6$  &  $\langle k\rangle = 8$ \\ 
        \hline
        $\Bar{e}_{sim}$ & 0.0137   & 0.0101     & 0.0077  & 0.0132    & 0.0096   &0.0078 \\
        $\Bar{e}_{our}$ & 0.0081   & 0.0072     & 0.0143  & 0.0011    & 0.0034   &0.0061 \\
        $\Bar{e}_{(sim,our)}$ & 0.0123   & 0.0043     & 0.0068  & 0.0085    & 0.0064   &0.0039 \\  
        \hline

    \end{tabular}
\label{tab:A_REF}

    \small \flushleft
     Note: $\Bar{e}_{sim}$ and $\Bar{e}_{our}$ represent the mean standard deviation of 100 simulated $L_{RS}$ and predicted $L_{RS}$ from our model. $\Bar{e}_{(sim,our)}$ is the difference between $p_i$ and $s_i$, $\Bar{e}_{(sim,our)} = \frac{\sum_{i=1}^L |p_i-s_i|}{L}$, where $p_i$ is the $i$th prediction of our model and $s_i$ is the $i$th simulation.

\hrule

\end{table}

\begin{table}[!htb]
    \caption{
    Mean standard deviations of simulation and prediction and the mean difference between them under high-edge-degree adaptive attack (HEDAA)
    }
    \centering
    \begin{tabular}{l|ccc|ccc}
        \hline
        \multirow{2}*{Error Types} &  \multicolumn{3}{c}{ER networks} & \multicolumn{3}{c}{BA networks} \\ 
        \cmidrule{2-7} 
        & $\langle k\rangle = 4$  &  $\langle k\rangle = 6$  &  $\langle k\rangle = 8$ & $\langle k\rangle = 4$  &  $\langle k\rangle = 6$  &  $\langle k\rangle = 8$ \\ 
        \hline
        $\Bar{e}_{sim}$ & 0.0065   & 0.0032     & 0.0021  & 0.0028    & 0.0039   &0.0026 \\
        $\Bar{e}_{our}$ & 0.0280   & 0.0095     & 0.0212  & 0.0420    & 0.0086   &0.0082 \\
        $\Bar{e}_{(sim,our)}$ & 0.0317   & 0.0129     & 0.0094  & 0.0517    & 0.0143   &0.0112 \\ 
        \hline

    \end{tabular}

    \small \flushleft
    Note: $\Bar{e}_{sim}$ and $\Bar{e}_{our}$ represent the mean standard deviation of 100 simulated $L_{RS}$ and predicted $L_{RS}$ from our model. $\Bar{e}_{(sim,our)}$ is the difference between $p_i$ and $s_i$, $\Bar{e}_{(sim,our)} = \frac{\sum_{i=1}^L |p_i-s_i|}{L}$, where $p_i$ is the $i$th prediction of our model and $s_i$ is the $i$th simulation.

\hrule

\label{tab:A_HEDAA}
\end{table}

\setcounter{figure}{0}

\begin{figure} [!htb]
\centering
\centerline{\includegraphics[width=12cm]{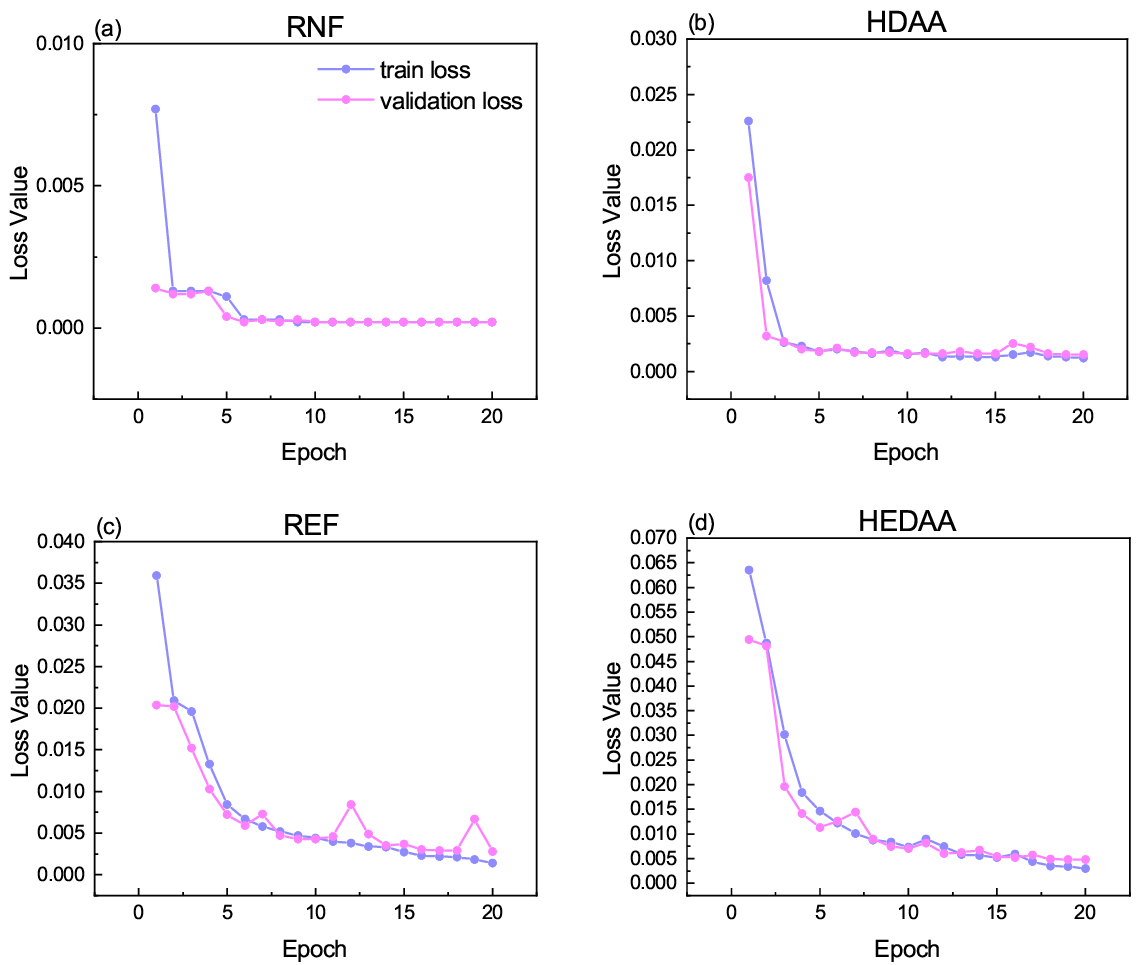}}
\caption{The loss value of four scenarios for the 20 epochs which is calculated by Equation ~\ref{eq:L2loss}. Train loss refers to the prediction error of the model on the training data, while validation loss represents the prediction error on the validation data. Both are utilized to assess the model's performance and generalization capability. And in each epoch, there are 4,800 training instances.}
\label{fig:loss}
\end{figure}

\begin{figure}[!htb]
    \centering
    \includegraphics[width=0.85\linewidth]{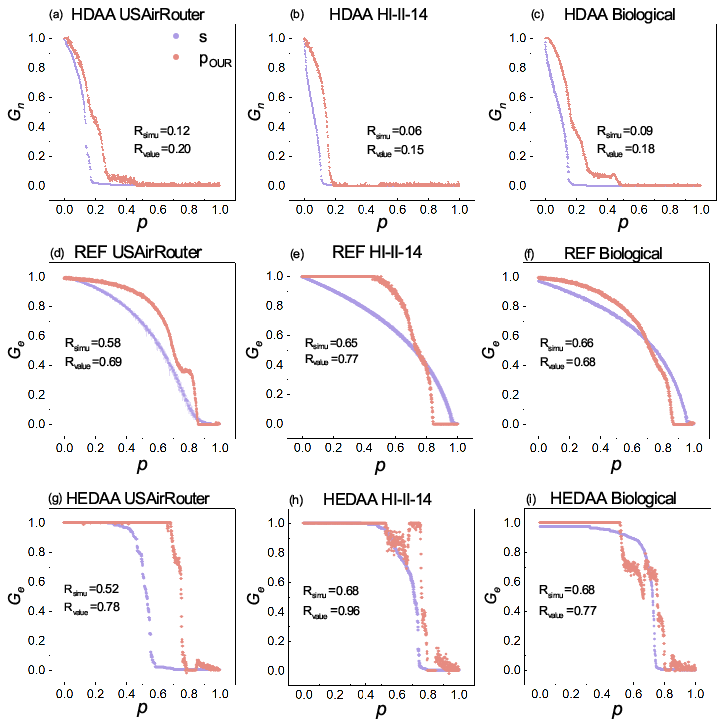}
    \caption{Attack curves and the robustness values on empirical networks under high-degree adaptive attack (HDAA), random edge failure (REF) and high-edge-degree adaptive attack (HEDAA). Here $p$ represents the proportion of removed nodes, $G_n$ and $G_e$ represent the relative size of the LCC respectively. The curve $s$ represents the simulated attack curve, while $p_\text{OUR}$ illustrates the predicted attack curve. $R_\text{simu}$ is the robustness value derived from the simulated attack curve, while $R_\text{value}$ is the robustness value directly predicted by our method.}
    \label{fig:empNet_3scenarios}
\end{figure}

\begin{table}[H]
\centering
\tabcolsep = 0.5cm
 \setlength{\abovecaptionskip}{0cm} 
        \setlength{\belowcaptionskip}{0.1cm}
\caption{Comparison of the time required for simulation (simu) and prediction (pre) of three empirical networks under four attack modes}

\begin{tabular}{cc|ccc}
\hline
\multicolumn{2}{c|}{Time(s)}                       & USAirRoutern & Biological & HI-II-14 \\ \hline
\multicolumn{1}{l|}{\multirow{2}{*}{RNF}}   & simu & 28           & 1113.2     & 241.6    \\ \cline{2-2}
\multicolumn{1}{l|}{}                       & pre  & 0.7          & 82         & 4.3      \\ \hline
\multicolumn{1}{l|}{\multirow{2}{*}{HDAA}}  & simu & 347.3        & 26334.9    & 4046.5   \\ \cline{2-2}
\multicolumn{1}{l|}{}                       & pre  & 0.6          & 72.2       & 4        \\ \hline
\multicolumn{1}{l|}{\multirow{2}{*}{REF}}   & simu & 3061         & 38714.9    & 7505.5   \\ \cline{2-2}
\multicolumn{1}{l|}{}                       & pre  & 0.6          & 79.1       & 4.1      \\ \hline
\multicolumn{1}{l|}{\multirow{2}{*}{HEDAA}} & simu & 19602        & 950400     & 29942
  \\ \cline{2-2}
\multicolumn{1}{l|}{}                       & pre  & 0.6          & 82.1       & 4.1      \\ \hline
\end{tabular}
\label{tab:A_Time}
\end{table}

\end{document}